\documentclass[10pt,twocolumn,letterpaper]{article}

\usepackage{iccv}
\usepackage{times}
\usepackage{epsfig}
\usepackage{graphicx}
\usepackage{amsmath}
\usepackage{amssymb}
\usepackage{multirow}
\usepackage{subfig}

\usepackage[breaklinks=true,bookmarks=false]{hyperref}

\iccvfinalcopy % *** Uncomment this line for the final submission

\def\iccvPaperID{****} % *** Enter the ICCV Paper ID here

\ificcvfinal\fi

\begin{document}

%%%%%%%%% TITLE
\title{Multi-Modality Cascaded Fusion Technology for Autonomous Driving}

\author{Hongwu Kuang, Xiaodong Liu, Jingwei Zhang, Zicheng Fang\\
Hikvision Research Institute, China\\
{\tt\small \{kuanghongwu,liuxiaodong9,zhangjingwei6,fangzicheng\}@hikvision.com}
}

\maketitle
\ificcvfinal\thispagestyle{empty}\fi

%%%%%%%%% ABSTRACT
\begin{abstract}
   Multi-modality fusion is the guarantee of the stability of autonomous driving systems. In this paper, we propose a general multi-modality cascaded fusion framework, exploiting the advantages of decision-level and feature-level fusion, utilizing target position, size, velocity, appearance and confidence to achieve accurate fusion results. In the fusion process, dynamic coordinate alignment(DCA) is conducted to reduce the error between sensors from different modalities. In addition, the calculation of affinity matrix is the core module of sensor fusion, we propose an affinity loss that improves the performance of deep affinity network(DAN). Last, the proposed step-by-step cascaded fusion framework is more interpretable and flexible compared to the end-to-end fusion methods. Extensive experiments on Nuscenes~\cite{caesar2019nuscenes} dataset show that our approach achieves the state-of-the-art performance.dataset show that our approach achieves the state-of-the-art performance.
\end{abstract}

%%%%%%%%% BODY TEXT
\section{Introduction}

With the development of sensors and environmental perception technology, how to integrate multi-view and multi-modality data through effective fusion technology to achieve accurate detection, classification and positioning has gain increasing attention in the field of autonomous driving.

Autonomous driving systems are generally equipped with sensors such as camera, LiDAR, millimeter wave radar and ultrasonic radar. The camera usually obtains richer information that has advantages in object classification, size estimation and angular positioning, but its velocity measurement and ranging accuracy is comparatively limited and can be further damaged by environmental factors such as light and weather. In contrast, radar systems have preferable ranging performance and are insensitive to environmental changes, but they provide relatively sparse information that is insufficient to sensing the object size and category. Sensors from single modality have inherent disadvantages that are not enough to handle the complex environment changes during the system running. So, it is necessary to fuse multi-modality sensors to obtain accurate and robust perception.

Multi-modality sensors fusion can be divided into decision-level, feature-level and data-level according to the input data~\cite{ramachandram2017deep}. Decision-level fusion regards the perception result of each sensor as input, and improves the reliability of the final decision by verifying decision results of multiple sensors mutually. However, feature-level and data-level fusion adopt abstracted multi-level features and source data of sensors as input, respectively, getting richer information while bringing more fusion difficulties. In recent years, there are many decision-level or feature-level based 3D detection, segmentation and tracking methods \cite{meyer2019sensor,madawy2019rgb,raffiee2019class,liang2019multi,xu2018multi,xu2018pointfusion}, bringing competitive results with a relatively simple form through an end-to-end data-driven approach. However, single-level fusion method is often unable to fully exploit the effective information of each sensor, and although the end-to-end approach simplifies the fusion process, it is not easy to interpret, which makes it difficult to locate and analyze when problems occur.

In this paper, we propose a general multi-modality cascaded fusion framework, combining decision-level and feature-level fusion methods, which can be further extended to arbitrary fusion of camera, LiDAR and radar with great interpretability. More specifically, the fusion process is along with multi-object tracking algorithm, which makes the fusion and association facilitate each other, simultaneously. Firstly, intra-frame fusion is performed and data from different sensors are processed. The intra-frame fusion uses a hierarchical method to associate targets with different confidence to form a main-sub format and enhances the fusion result by adopting dynamic coordinate alignment of multi-modality sensors. Secondly, inter-frame fusion is executed when intra-frame is completed, which extends the tracking lists by matching targets in adjacent frames and can handle single sensor failure. Finally, we propose an affinity loss that boosts the performance of the deep affinity networks, which used in intra-frame and inter-frame instead of artificial strategy, and improves the results of association. Extensive experiments conducted on NuScenes dataset demonstrate the effectiveness of the proposed framework.

In summary, the contributions of this paper are as follows:
\begin{itemize}
\item We propose a multi-modality cascaded fusion framework, which makes full use of multi-modality sensor information, improving robustness and accuracy with great interpretability.
\item We propose a dynamic coordinate alignment method of sensors from different modalities that reduces the problems of heterogeneous data fusion.
\item We propose an affinity loss used in training deep affinity estimation networks, which improves the performance of multi-modality data association.
\end{itemize}

\section{Related Work}

\subsection{Multi-Modality Fusion}

Multi-modality can be divided into decision-level~\cite{kahou2016emonets}, feature-level~\cite{lenz2015deep} and data-level~\cite{valada2016deep}. In autonomous driving, thanks to the arise of deep learning, not only single sensor research has made great progress \cite{li2019gs3d,lu2019l3,li2019stereo,qi2017pointnet}, such as camera, LiDAR, multi-modality fusion has also gain increasing attention \cite{liang2019multi,xu2018pointfusion,chen2017multi,rashed2019fusemodnet}, these methods usually adopt decision-level or feature-level fusion. \cite{liang2019multi} proposes a multi-task network, which integrates vision and LiDAR information, achieving ground estimation, 3D detection, 2D detection and depth estimation, simultaneously. \cite{xu2018pointfusion} adopts PointNet~\cite{qi2017pointnet} and ResNet~\cite{he2016deep} to extract point cloud and image feature, respectively, then fuses them to get object 3D bounding boxes. \cite{chen2017multi}collects the RGB image, the front view and top view of the LiDAR, and obtains 3D bounding box by feature-level fusion. \cite{rashed2019fusemodnet} proposes a real-time moving target detection network, which captures the motion information of vision and LiDAR and exploits the characteristics of LiDAR that are not affected by light to compensate the camera failure in low-light condition. The framework proposed in this paper combines decision-level and feature-level fusion and achieves robust and effective multi-modality fusion.

\subsection{Multi-Object Tracking}

Multi-object tracking is a core task in autonomous driving, because the control and decision of the vehicle depend on the surrounding environment, such as the movement of pedestrians and other cars. Many multi-object tracking algorithm only consider visual input, which concern more about the task itself, such as concentrating on different multi-object tracking framework~\cite{bergmann2019tracking}, solving association problem by graph optimization~\cite{kim2015multiple}, or replacing the traditional association process with an end-to-end method~\cite{sun2019deep}. However, visual based multi-object tracking can be easily invalid under extreme light conditions, and is infeasible to obtain accurate target distance and velocity. \cite{held2013precision,mitzel2012taking} use information more than a single RGB image and get accurate inter-frame transformations in the tracking process. \cite{zhang2019robust} proposes a multi-modality multi-object tracking framework, which obtains the tracking lists by directly fusing the detection results of visual and LiDAR in a deep network, but the fusion between visual and LiDAR is simple and lack of mutual promotion of different sensors in the association process. However, the framework proposed in this paper forms a main-sub format that reserves multi-modality sensors information, which reinforces the hierarchical association result of multi-modality data and is more robust to single sensor failure.

\section{Methodology}

In this paper, we propose a multi-modality cascaded fusion framework, investigating the advantages of the decision-level and feature-level fusion, achieving the accuracy and stability while keeping the interpretability.
 
\begin{figure}
	\begin{center}
		\includegraphics[width=0.45\textwidth]{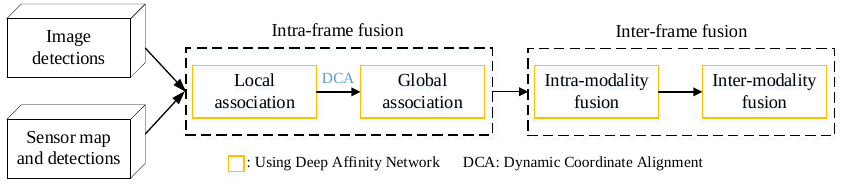}
	\end{center}
	\caption{The multi-modality cascaded fusion framework.}
	\label{fig:framework}
\end{figure}

As shown in Figure~\ref{fig:framework},the proposed cascaded framework is composed of two parts, namely, the intra-frame fusion and the inter-frame fusion. Firstly, the intra-frame fusion module fuses the vision and radar detections within each frame, producing fused detections in the main-sub format. Secondly, the inter-frame fusion performs association between the tracklets at time $t-1$ and the fused detections at time $t$, and generates accurate object trajectories.

In intra-frame fusion, the local and the global association are progressive not only in order, but also in functions. By using dynamic coordinate alignment, multi-modality data can be mapped into a same coordinate, generating more reliable feature similarity evaluation. In addition, because the calculation of similarity is the core of data association, a deep affinity network is designed to realize accurate similarity evaluation throughout the entire process of the intra-frame and inter-frame fusion.

\subsection{Intra-Frame Fusion}

The intra-frame fusion is the first step of proposed cascaded fusion framework, which aims at fusing the multi-modality detections within each frame. After intra-frame fusion, we can obtain a series of fused detections in the main-sub format, which contain information from one or more modalities. Specifically, it performs by two sequential steps: the local association and the global association.

Let $V^T=\{v_m^T|m=1,...,M\}$ and $R^T=\{r_n^T|n=1,...,N\}$ denote the vision detections and radar detections at time $T$, respectively. All the vision and radar detections are distinguished into high-confidence and low-confidence, namely, $VH^T$, $VL^T$, $RH^T$ and $RL^T$ according to the detection confidence and the corresponding threshold $\delta_V$ and $\delta_R$. During the intra-frame fusion, the high-confidence vision and radar detections are collected and sent into the local association. The association cost matrix between $VH^T$ and $RH^T$ is calculated as follows:

\begin{equation}
C=F(\tau(\varphi(VH^T)),\tau(\varphi(RH^T))),
\label{eq1}
\end{equation}
where $\varphi$ denotes the extracted common feature for vision or radar object, including range, angle, velocity and confidence, $\tau$ fuses the feature vectors of $VH^T$ and $RH^T$, $F$ is the deep network for similarity evaluation, which will be introduced in section \ref{sec:DAN}.

By employing Hungarian algorithm~\cite{munkres1957algorithms} on the cost matrix $C$, we can acquire an assignment matrix $H$ with elements 0 and 1.For each vision-radar pair satisfying assignment $H_{ij}=1$ and similarity $C_{ij}$ greater than threshold $\delta_{local}$, the corresponding vision and radar are matched, forming a main-sub (vision-radar) detection $V_iR_j$. While the remainder vision and radar detections in $VH^T$ and $RH^T$ are then added to $VL^T$ and $RL^T$, respectively, competing in the successive association.

Note that, the local association is generally reliable due to the high object confidence, hence we utilize the matching results to dynamic align sensor coordinate to reduce the mapping error of heterogeneous and facilitate successive fusion.

In order to realize further fusion, the global association is conducted after the local association, which consider all the low-confidence detections $VL^T$ and $RL^T$, as well as the unassigned ones in $VH^T$ and $RH^T$. In addition, due to the antenna azimuth resolution limitation, radar sensor is often unable to distinguish different targets in dense scenes. To cope with that, we adjust the assign strategy in global association, which allows one-to-many assignment, i.e., a low-confidence radar detection could match multiple vision detections when the similarities are high enough. The calculation of cost matrix is the same as local association

After intra-frame fusion, we can obtain three types of fusion results, namely, $VR$, $V$ and $R$ as the input of the following inter-frame fusion.

\subsection{Dynamic Coordinate Alignment}

Generally, the similarity computation between heterogeneous sensor data relies on their common features. For example, in $VR$ systems, these can be the object range, velocity, angle and confidence under a specifically defined geometry space. However, due to the perception principle difference, although the geometry relationship of camera and radar sensor can be calibrated in advance, it may change inevitably during driving, leading to a non-equivalent common feature, e.g., the object range, and finally impacts on associated results. Therefore, a dynamic sensor coordinate alignment is critical.

or vision system with a monocular camera, the range feature can be extracted by two conventional means: the size ranging and the trigonometric ranging. The size ranging uses the proportionate information between the image and the real physical size according to the pinhole camera geometry. Such method relies on the accuracy of a priori object size, which is usually hard to obtain in practice. The trigonometric ranging assumes interested objects are in the same plane with road surface, its performance mainly depends on the accuracy of given vanishing horizontal line.

\begin{figure}[t]
	\begin{center}
		\includegraphics[width=1.0\linewidth]{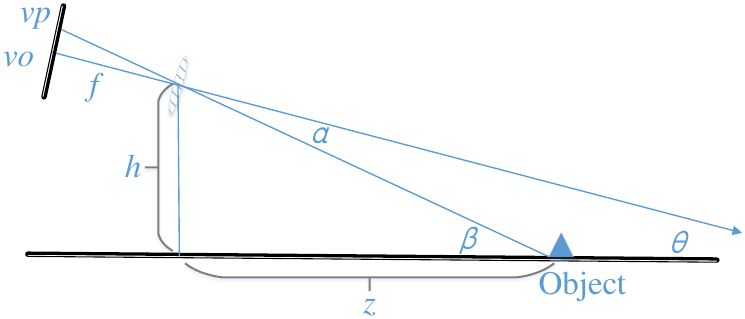}
	\end{center}
	\caption{The trigonometric ranging model.}
	\label{fig:DCA}
\end{figure}

In order to realize dynamic coordinate alignment, we propose a vanishing horizontal line compensation approach by using the radar ranging information of local association results or the historical image ranging results. The trigonometric ranging model is shown in Figure \ref{fig:DCA}.

Assume that the road is planar, the camera optical axis is parallel to the road surface with camera pitch angle $\theta=0$, thus the object range can be calculated as:
\begin{equation}
Z = \frac{h}{\tan(\beta)} = \frac{h}{\tan(\alpha+\theta)} = \frac{h}{\tan(\alpha)},
\label{eq2}
\end{equation}
where $h$ is camera height, $\alpha$ is the angle between optical axis and the line formed by optical center and the ranging point.

So, we can get $\alpha=\arctan\frac{vp-vo}{f}$, where $vp$ and $vo$ denote the image height pix of object bottom center and the optical center respectively, then we can obtain the distance $Z$ using Equation \ref{eq2}. However, the pitch angle $\theta$ might change during driving as road is not always flat, which consequently leads to an inaccurate value of $Z$. It is worth noting that radar ranging is generally accurate and stable due to the sensor characteristic, which can be used to compensate the camera pitch angle. Specifically, let $Z'$ denote the radar ranging of a reliable $VR$ pair from local association, such that we have 

\begin{equation}
\alpha + \theta=\arctan\frac{h}{Z'},
\label{eq3}
\end{equation}

\begin{equation}
\theta=\arctan\frac{h}{Z'}-\arctan\frac{vp-vo}{f}.
\label{eq4}
\end{equation}

Hence, by using multiple local association $VR$ pairs, the camera pitch angle can update immediately, improving the performance of global association.

\subsection{Deep Affinity Network}
\label{sec:DAN}

\begin{figure}[t]
	\begin{center}
		\includegraphics{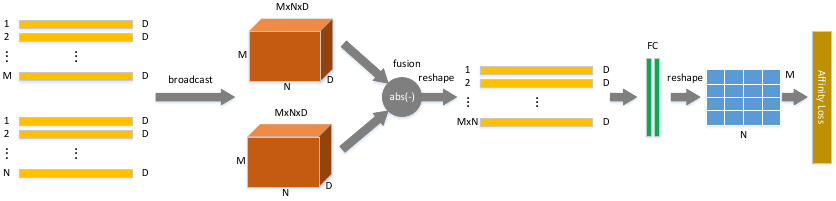}
	\end{center}
	\caption{The architecture of deep affinity network.}
	\label{fig:DAN}
\end{figure}

As the core of the association, the deep similarity computation network proposed in this paper is shown in Figure \ref{fig:DAN}.

For convinience, denoting $A=\{A_i \in R^D|i=1,...,M\}$ and $B=\{B_j \in R^D|j=1,...,N\}$ as outputs of sensor $A$ and $B$, respectively, where $D$ is the dim of extracted feature vectors, $M$ and $N$ are the corresponding number of detections. The feature vectors forwarded in this paper contain the range, angle, velocity, size, confidence and even abstracted feature extracted from raw image and radar power map. The $M\times D$ and $N\times D$ inputs are converted to the $M\times N\times D$ feature map by broadcast. Unlike traditional similarity computation by Euclidean or cosine distance, we instead use a deep affinity network. Let $I=\{I_{ij}=|A_i-B_j||i=1,...,M;j=1,...,N\}$ denotes the fused feature vectors of sensors $A$ and $B$. Reshaping the feature vector $I\in R^{(M\times N)\times D}$ as the input, we train the multi-layer fully connected network $F$ to predict the final affinity matrix $C\in R^{M\times N}$.

The label of the affinity matrix is defined as follows:
\begin{equation}
G_{ij}=\left\{
\begin{array}{ll}
0 & \mbox{$A_i$ and $B_j$ are not the same object}\\
1 & \mbox{$A_i$ and $B_j$ are the same object}
\end{array},
\right.
\end{equation}

We designed two loss functions including mask loss and affinity loss in this paper for training network.

\textbf{Mask loss.} Mask loss is simple and easily understandable because the cost is directly compared with the label of the affinity matrix. It is defined as Equation ~\ref{eq:mask_loss}.

\begin{equation}
\begin{split}
L(A,B)=\frac{1}{M*N}\sum_{i=1,...,M;j=1,...,N} |C_{ij}-G_{ij}|
\end{split}
\label{eq:mask_loss}
\end{equation}

\textbf{Affinity loss.} In fact, we use the Hungarian algorithm~\cite{munkres1957algorithms} to acquire matching pairs, instead of fitting the network output to the label, which is often difficult. Therefore, we just need to train the network to meet the following condition:

\begin{equation}
\begin{split}
C_{ij}>C_{i*} \ and \ C_{ij}>C_{*j} \\ s.t. \quad G_{ij}=1,G_{i*}=0 \ and\  G_{*j}=0.
\end{split}
\end{equation}

In the other word, if $A_i$ and $B_j$ are the same object, the corresponding cost $C_{ij}$ should be the maximum of the $i$-th row and $j$-th colume of the affinity matrix $C$.

Based on the above analysis, we propose an affinity loss function as Equation ~\ref{eq:affinity_loss}:

\begin{equation}
\begin{split}
L(A,B)=\sum_{i=1,...,M;j=1,...,N;G_{ij}=1} l(C_{ij}, C),\\
l(C_{ij}, C)=\sum_{k=1,...,N;G_{ik}\neq1} \max(0,C_{ik}-C_{ij}+m) \\ +\sum_{p=1,...,M;G_{pj}\neq1} \max(0,C_{pj}-C_{ij}+m),
\end{split}
\label{eq:affinity_loss}
\end{equation}
where $m$ denotes the margin between the positive and negative samples. The affinity loss encourages the network to fit better when $m$ is larger. Compared with mask loss, the affinity loss makes the network to converge better and faster.

\subsection{Inter-Frame Fusion}

The aim of inter-frame fusion is to associate the tracklets at time $t-1$ and the intra-frame fusion results at time $t$, which includes homologous data association and heterogeneous data association. The inputs of inter-frame fusion are objects $VR$, $V$ and $R$ produced by intra-frame fusion.

\begin{figure}
	\begin{center}
		\includegraphics[width=1.0\linewidth]{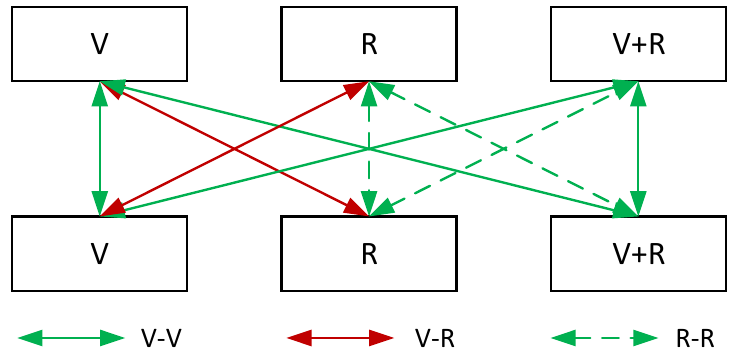}
	\end{center}
	\caption{Figure 4. 	The diagram of inter-frame fusion, first row: tracklets at time $ t-1 $, second row: fused detections at time $ t $.}
	\label{fig:VR}
\end{figure}

As shown in the figure \ref{fig:VR}, the inter-frame fusion contains multiple strategies for different matching mode, which is 9 in this paper.

There are 7 types of the homologous data association and 2 types of the heterogeneous data association. The homologous data association is more reliable because richer common features and naturally identical coordinate. Therefore, the homologous data association is firstly conducted to reduce the interference of other data. For example, when two $RV$ objects are computed for similarity, image similarity $V$-$V$ and radar similarity $R$-$R$ are computed separately and then combined to the ultimate similarity. When computing $RV$ object and image object $V$, only image similarity $V$-$V$ need to be computed. The heterogeneous data association, for example $V$-$R$ and $R$-$V$, needs to be considered in the same dimension. Therefore, we first perform the coordinate conversion between image and radar data, and then extract common features such as the range, angle, velocity, confidence for similarity computation.

The stable and reliable tracking lists are extended after the hierarchical association, and multiple matching strategies make the fusion results insensitive to single modality sensor failure.

\section{Experiments}

\subsection{Dataset}
\label{dataset}

We evaluate the proposed multi-modality cascaded fusion technology on NuScenes dataset. NuScenes is a well-known public dataset including LiDAR, radar, camera and GPS unit for autonomous driving. About 1000 scenes with 3D bounding boxes are available, covering various weather and road conditions. There are 6 different cameras and 5 radars on the test vehicle. However, in this work, we only use the front camera and the front radar.

We randomly split the dataset into train, validation and test sets with 275, 277 and 275 scenes, respectively. We combine multiple vehicles, such as car, truck, bus and so on, into one single class car and get 37907 objects in the test set.

\begin{figure}
	\begin{center}
		\includegraphics[width=1.0\linewidth]{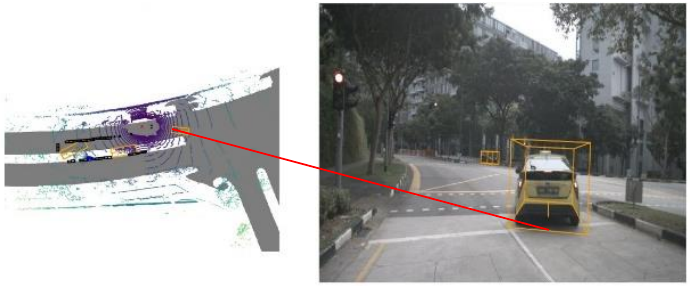}
	\end{center}
	\caption{Our method to generate the ground true distances.}
	\label{fig:distanceGT}
\end{figure}

However, the dataset lacks distance annotation for each object. \cite{zhu2019learning} proposed a method to obtain the ground truth distance and the keypoint of each object. They extracted the depth value of the $n$-th sorted laser point clouds as the distance of the object. It is unreasonable because they were more concerned about using the keypoint to calculate the projection loss for enhanced model than using it as the true value. To solve the problem, we propose a new method to generate the ground true distances for the test set, as shown in Figure \ref{fig:distanceGT}.

\begin{figure}
	\begin{center}
		\includegraphics[width=1.0\linewidth]{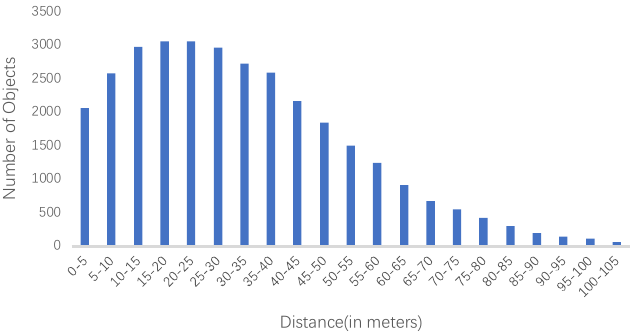}
	\end{center}
	\caption{Distribution of the ground true distances of objects in test set.}
	\label{fig:distance-dist}
\end{figure}

We project the 3D bounding boxes provided by the dataset into bird’s-eye view coordinates. Two corners closest 
to the test vehicle are extracted and the midpoint is then calculated. The distance between the midpoint and the bumper of the test vehicle is the ground true distance for the object. The distribution of the distances in the test set range from 0 to 105m, most within 5 to 40m, as shown in Figure \ref{fig:distance-dist}.

\subsection{Implementation Details}
The experiments include two parts, public comparison and ranging accuracy comparison.

\subsubsection{Public comparison}
For fairness, we evaluate our method on NuScenes mini dataset using classic evaluation metrics for depth prediction, including threshold proportion ( ), absolute relative difference (Abs Rel), squared relative difference (Squa Rel), root of mean squared errors (RMSE) and root of mean squared errors (RMSElog). And the ground truth distance is obtained the same way as ~\cite{zhu2019learning}.

In this part, we show the effectiveness of the proposed cascaded fusion framework without DCA and DAN. The distance of the image object is obtained by our monocular ranging model with triangulation and scale ranging method. Besides, the camera and radar need to be calibrated in advance and the similarity computation is designed by artificial.

The similarity computation is as follows:

\begin{center}
\begin{equation}
\begin{split}
C_{ij}=\omega_r*F_r+\omega_a*F_a+\omega_v*F_v,\\
F_r = \frac{|r_v-r_r|}{thr_r},F_a = \frac{|a_v-a_r|}{thr_a},F_v = \frac{|v_v-v_r|}{thr_v},
\end{split}
\label{eq:simi_arti}
\end{equation}
\end{center}
where $F_r$ denotes the similarity feature of the range, $F_a$ denotes the similarity feature of the angle and $F_v$ denotes the similarity feature of the velocity. $\omega_r$, $\omega_a$ and $\omega_v$ denote the weight of corresponding feature, which are obtained by policy searching. $r_v$, $a_v$ and $v_v$ denote the range, angle and velocity of vision, while $r_r$, $a_r$ and $v_r$ denote the range, angle and velocity of radar. $thr_r$, $thr_a$ and $thr_v$ are thresholds used for normalization.

The affinity matrix is obtained by the traditional method above for all radar and image objects at the same time. Then the matching results are acquired by the Hungarian algorithm Public comparison on NuScenes mini dataset and the range and velocity of the vision object are updated by the associated radar object.

\subsubsection{Ranging accuracy comparison}
We conduct extensive experiments on the split test dataset to evaluate the proposed DCA, DAN and cascaded fusion framework using ranging accuracy. Ranging accuracy is the proportion of the number of correct ranging objects with ranging error within $10\%$ to the ground truth, which is obtained by our method described in section \ref{dataset}.

All the similarities are computed by the DAN with two fully connected (FC) layers and a sigmoid activation, which makes the output varies from 0 to 1.

The DAN is trained using SGD optimizer with learning rate of 0.001 and batch size of 1.

\subsection{Results}
\subsubsection{Public comparison}
We compare our method with support vector regression (SVR) ~\cite{gokcce2015vision}, inverse perspective 
mapping algorithm(IPM) ~\cite{tuohy2010distance} and an enhanced model with a keypoint regression(EMWK) ~\cite{zhu2019learning}. We first test the proposed framework with camera and radar, but without DCA and DAN, namely \textit{Baseline}. Then we reduce the radar inputs to test the robust of the framework, namely \textit{Baseline-R}.

The results are shown in Table \ref{table:pub}. Obviously, our method \textit{Baseline-R} surpasses the SVR, IPM and EMWK with a large margin, which demonstrate the superiority and robustness of the framework. Furthermore, after adding radar input, \textit{Baseline} is better than \textit{Baseline-R}, which proves that radar contributes to the ranging performance and means the proposed cascaded fusion framework is effective.

\begin{table*}
	\centering
	\begin{tabular}{r|ccc|cccc}
		\hline
		& $\delta_1\uparrow$ & $\delta_2\uparrow$ & $\delta_3\uparrow$ & $AR\downarrow$ & $SR\downarrow$  & $RMSE\downarrow$ & $RMSE_{log}\downarrow$ \\
		\hline
		\textit{SVR~\cite{gokcce2015vision}}        & 0.308  & 0.652  & 0.833  & 0.504 & 13.197 & 18.480  & 0.846     \\
		\textit{IPM~\cite{tuohy2010distance}}        & 0.441  & 0.772  & 0.875  & 1.498 & 1979.375 & 249.849 & 0.926     \\
		\textit{EMWK~\cite{zhu2019learning}}       & 0.535  & 0.863  & 0.959  & 0.270 & 3.046    & 10.511  & 0.313     \\
		\textit{Baseline-R} & 0.776  & 0.937  & 0.982  & 0.163 & 2.099    & 9.979   & 0.208     \\
		\textit{Baseline}   & \textbf{0.811}  & \textbf{0.950}  & \textbf{0.988} & \textbf{0.133} & \textbf{2.032}    & \textbf{9.870}   & \textbf{0.202}    \\
		\hline
	\end{tabular}
	\caption{Public comparison on NuScenes mini dataset.$\uparrow$ means higher is better, $\downarrow$ means lower is better.}
	\label{table:pub}
\end{table*}

\begin{figure}[ht]
	\centering
	\begin{minipage}[b]{1.0\linewidth}
		\subfloat[Urban road]{
			\begin{minipage}[b]{0.45\linewidth}
				\centering
				\includegraphics[width=\linewidth]{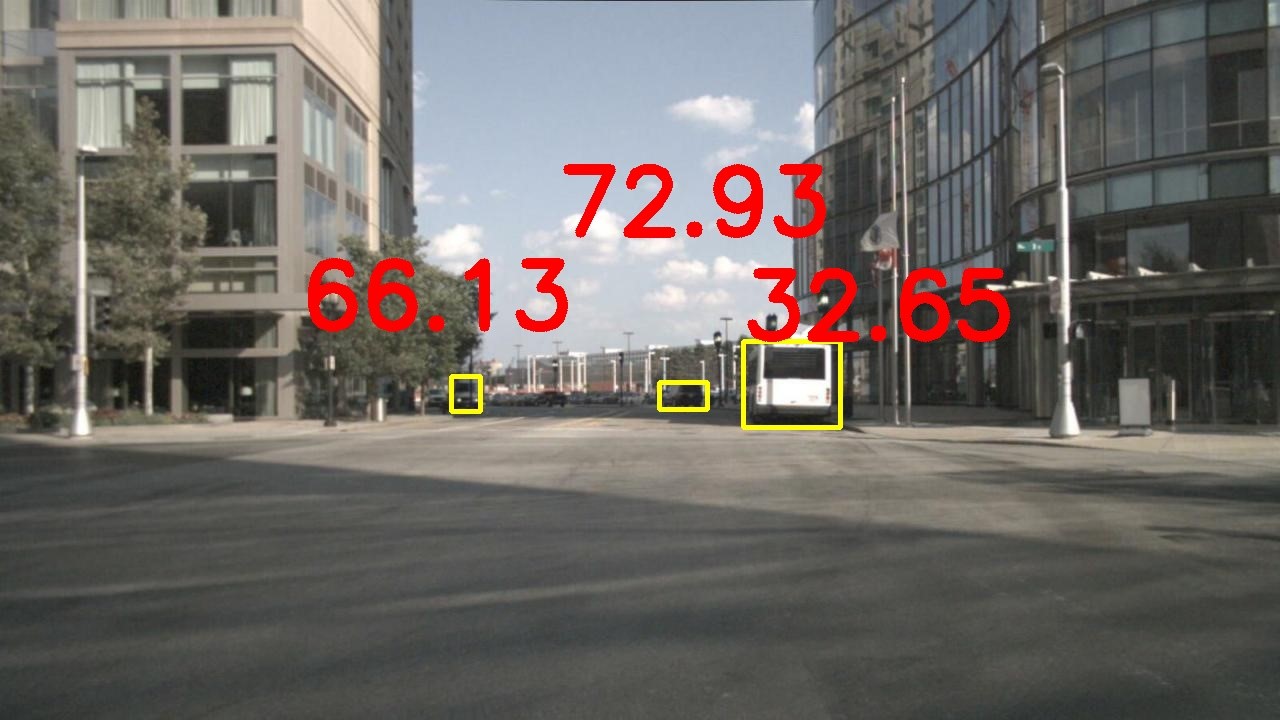}\vspace{8pt}
				\includegraphics[width=\linewidth]{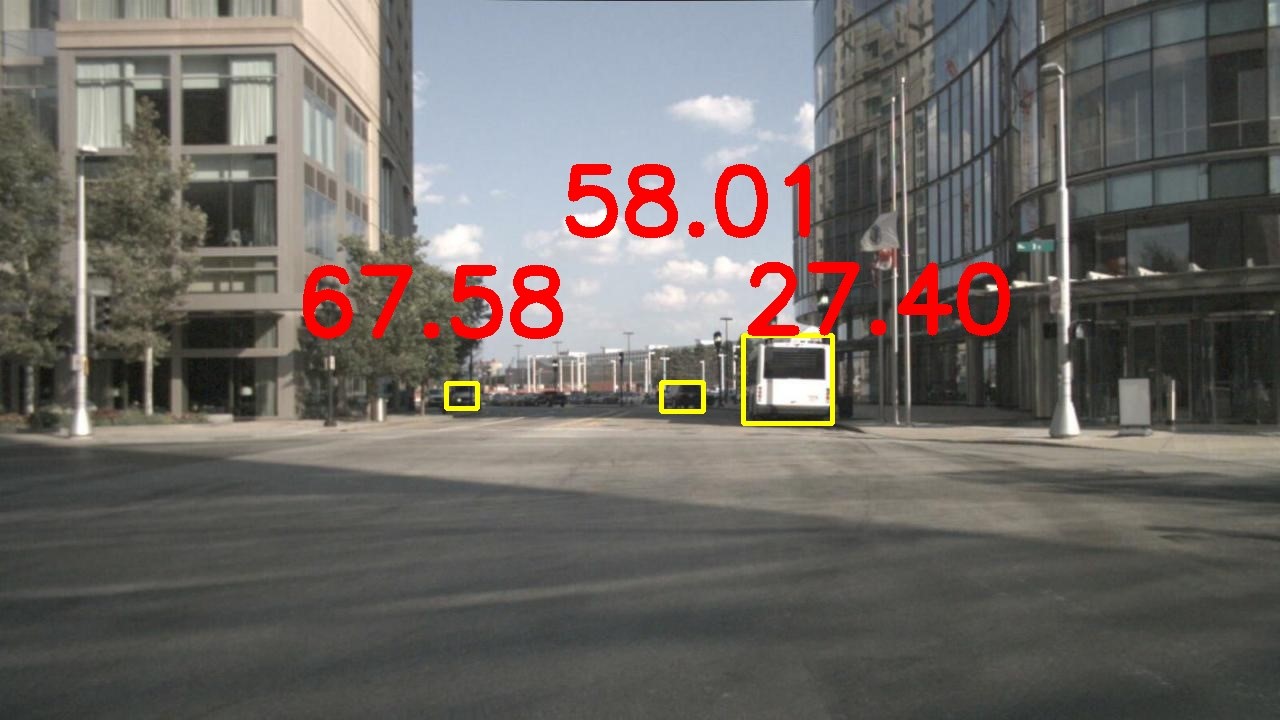}\vspace{8pt}
				\includegraphics[width=\linewidth]{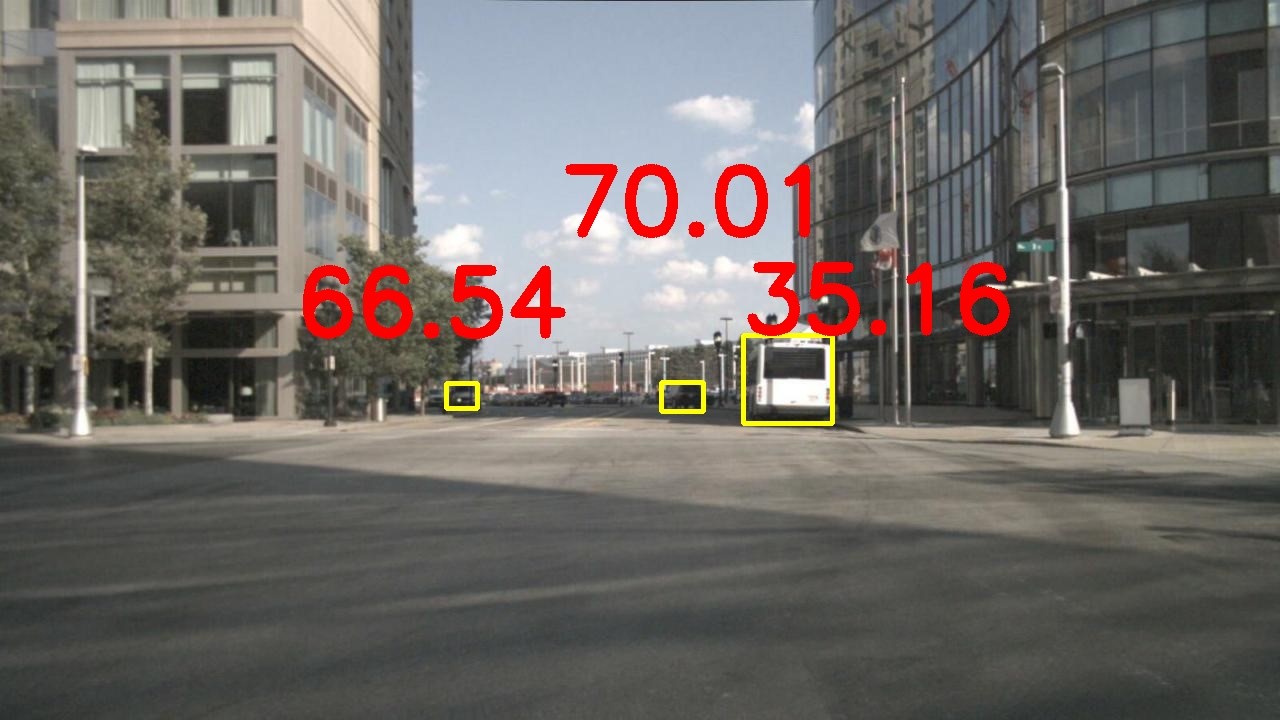}\vspace{8pt}
			\end{minipage}
		}
		\hfill
		\subfloat[Rough road]{
			\begin{minipage}[b]{0.45\linewidth}
				\centering
				\includegraphics[width=\linewidth]{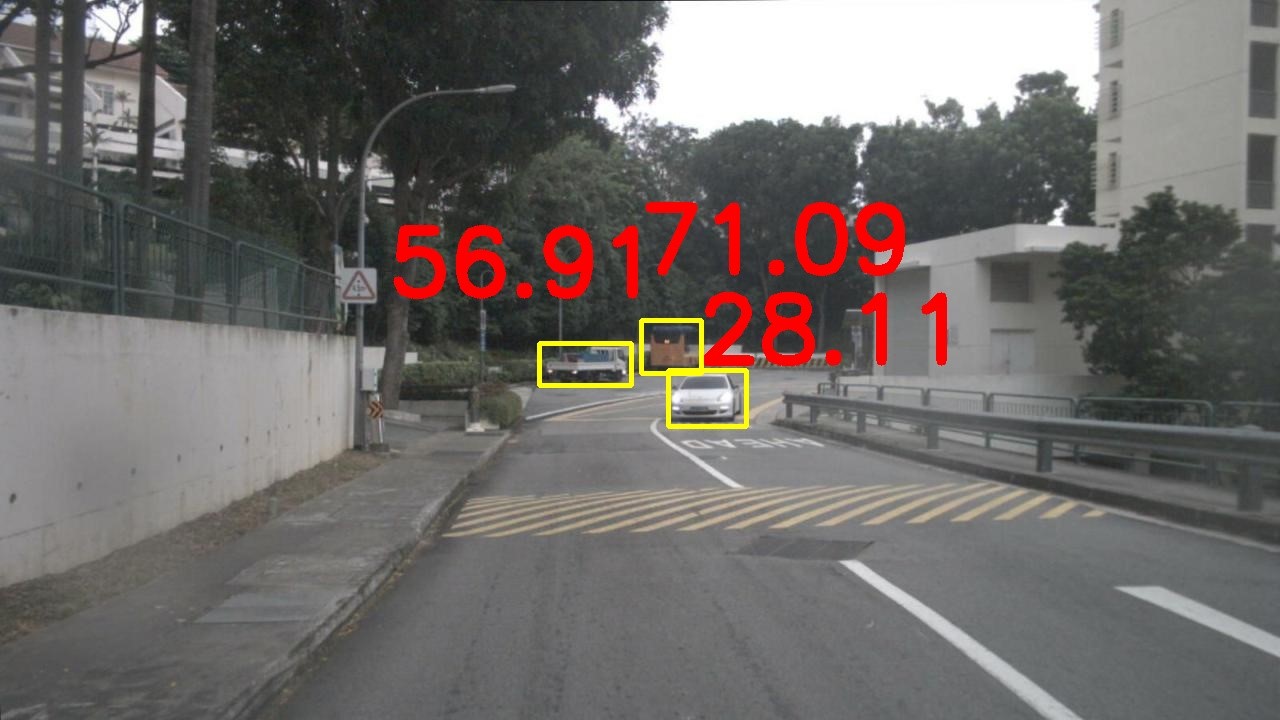}\vspace{8pt}
				\includegraphics[width=\linewidth]{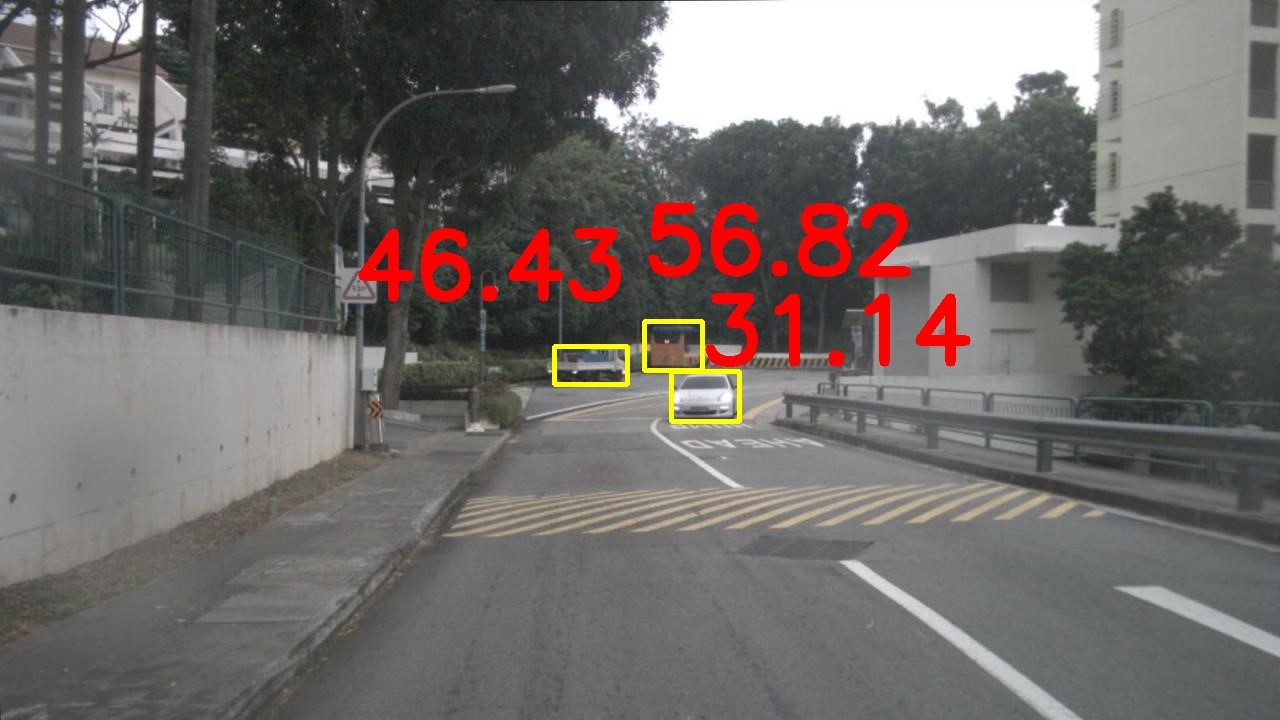}\vspace{8pt}
				\includegraphics[width=\linewidth]{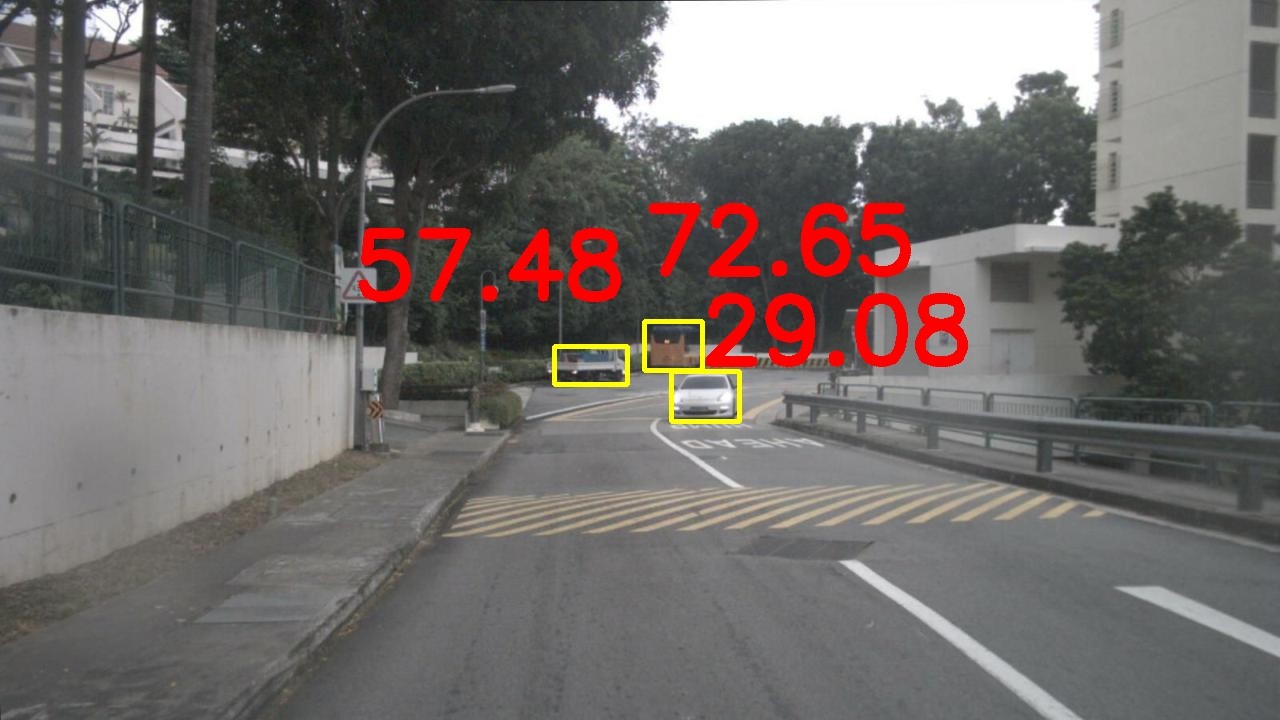}\vspace{8pt}
			\end{minipage}
		}
		\hfill
		\subfloat[Night]{
			\begin{minipage}[b]{0.45\linewidth}
				\centering
				\includegraphics[width=\linewidth]{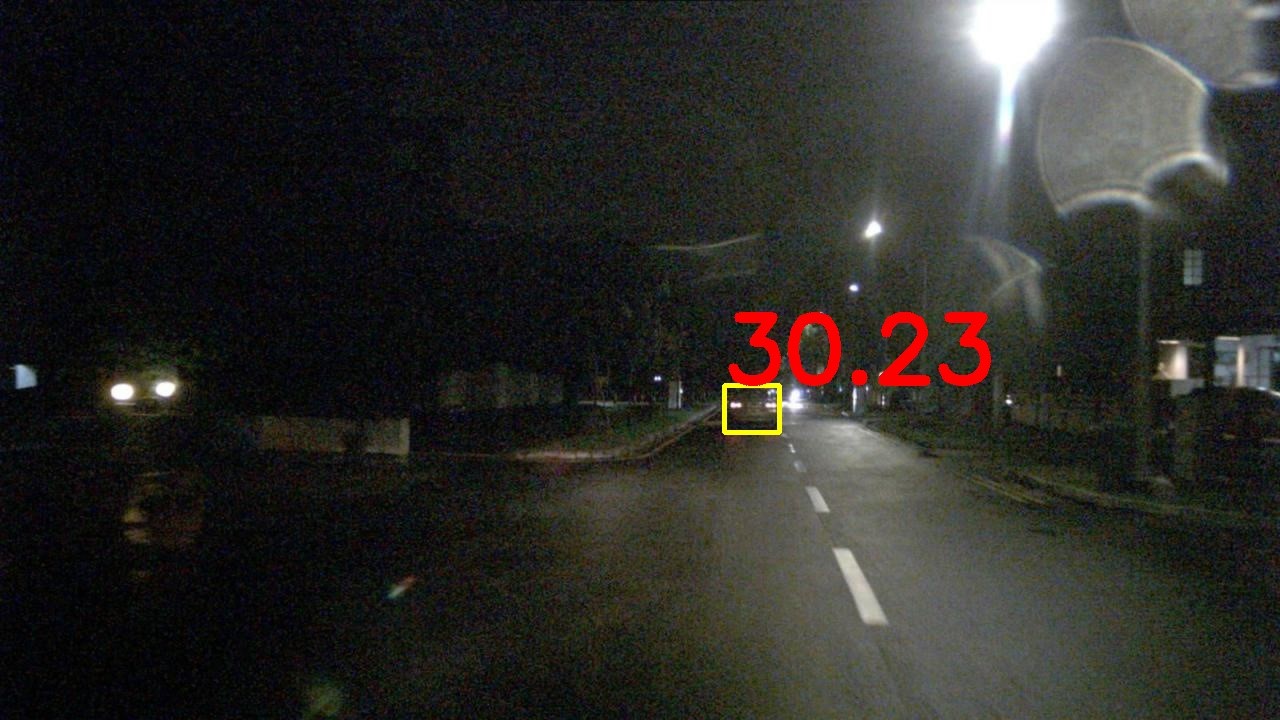}\vspace{8pt}
				\includegraphics[width=\linewidth]{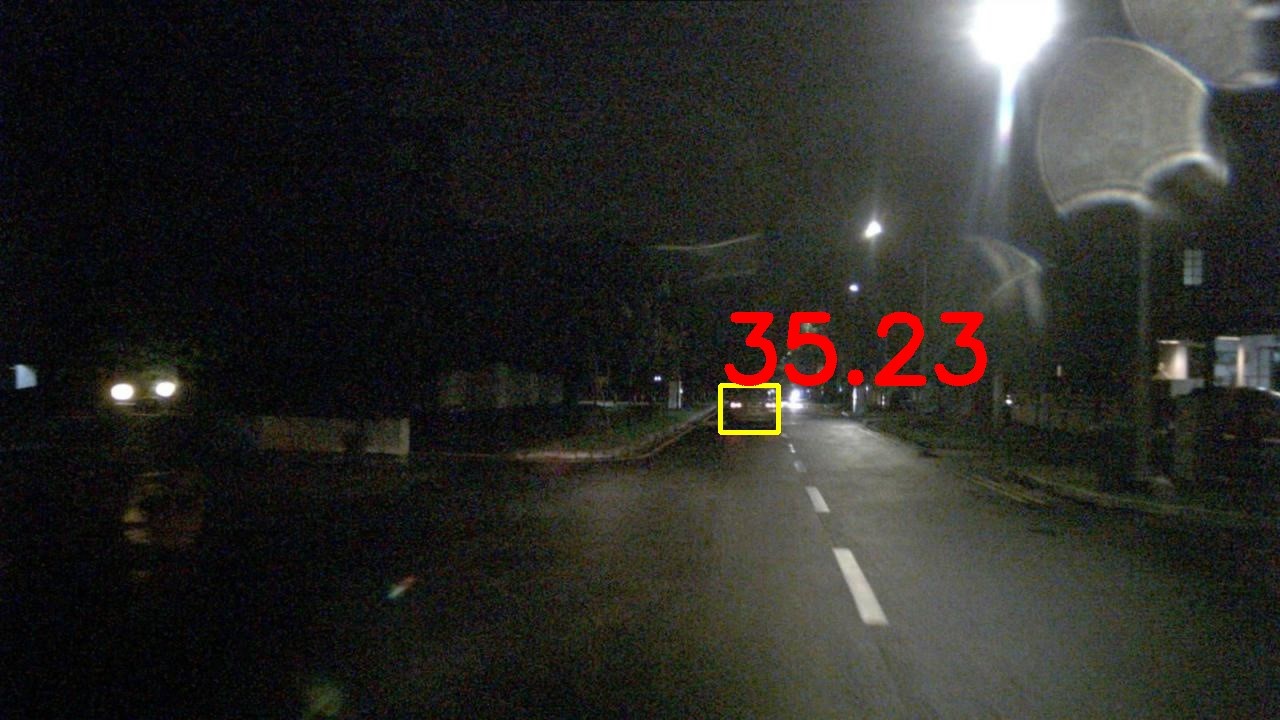}\vspace{8pt}
				\includegraphics[width=\linewidth]{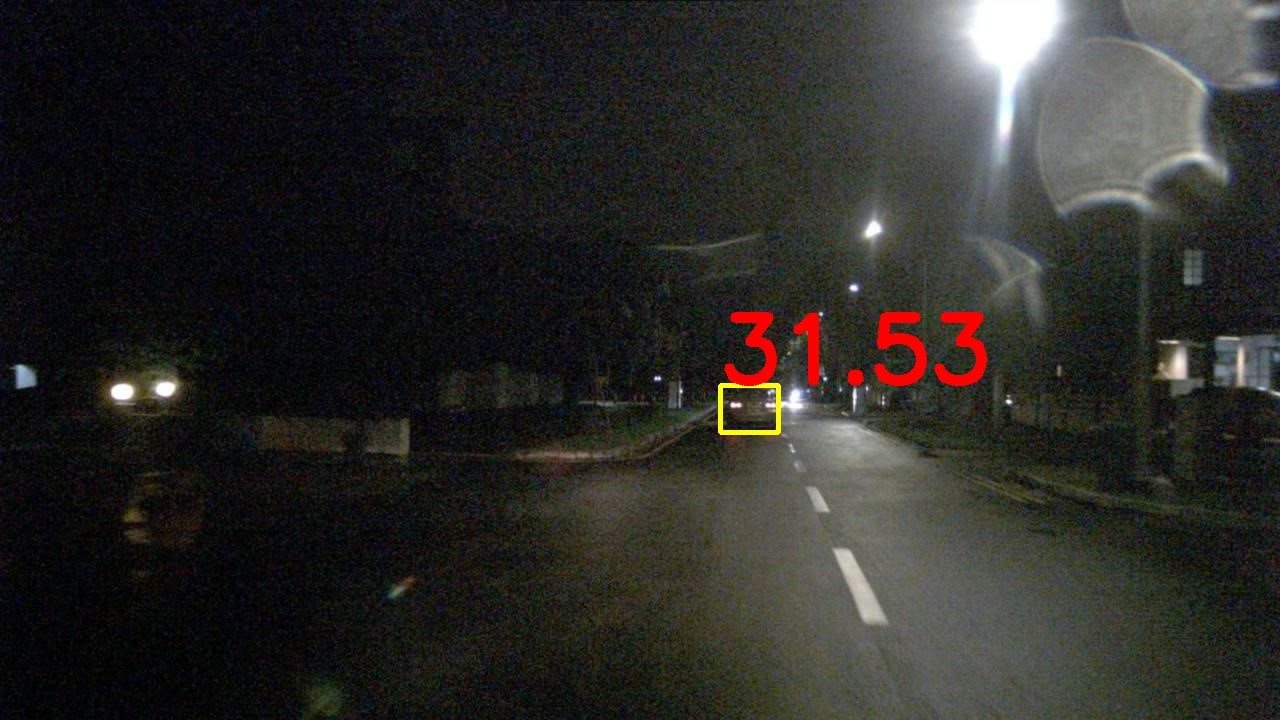}\vspace{8pt}
			\end{minipage}
		}
		\hfill
		\subfloat[Rain]{
			\begin{minipage}[b]{0.45\linewidth}
				\centering
				\includegraphics[width=\linewidth]{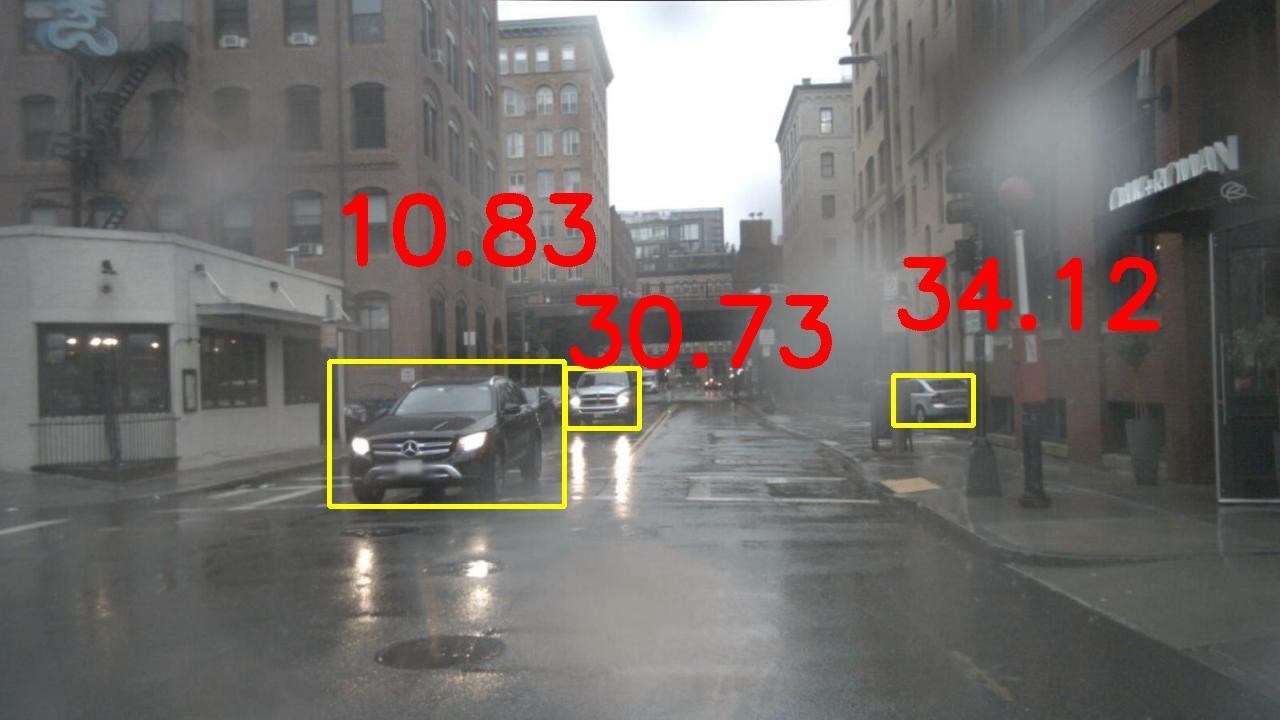}\vspace{8pt}
				\includegraphics[width=\linewidth]{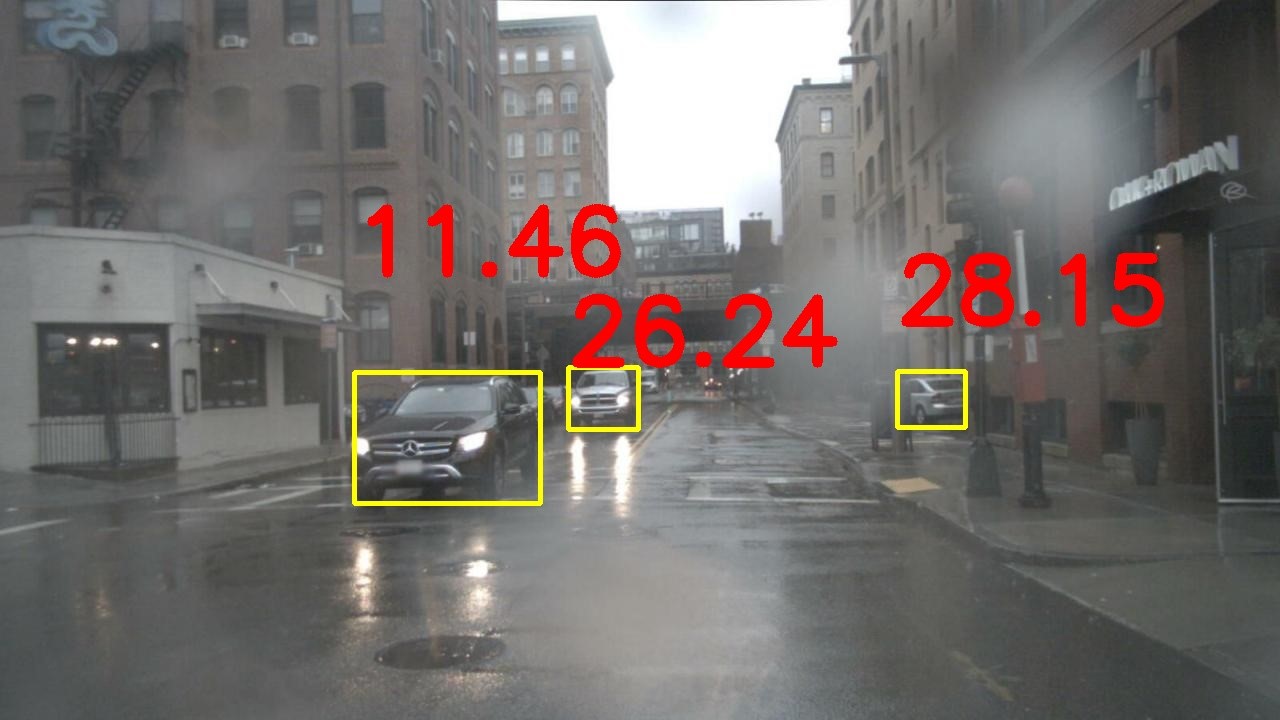}\vspace{8pt}
				\includegraphics[width=\linewidth]{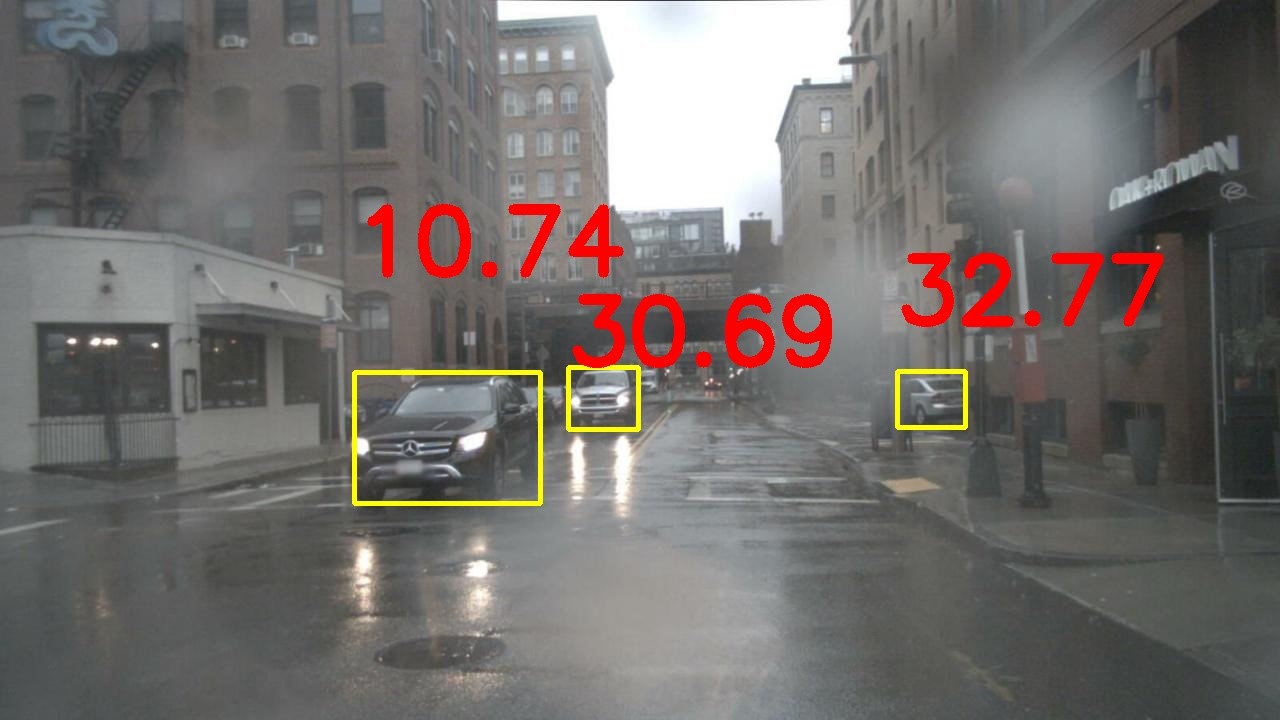}\vspace{8pt}
			\end{minipage}
		}
	\end{minipage}
	\vfill
	\caption{Examples of the estimated distance. Top row: ground truth, middle row: \textit{Baseline - R}, bottom row: \textit{Baseline + DCA + DAN-al}.}
	\label{exp:qualitative}
\end{figure}

\subsubsection{Ranging accuracy comparison}

In this part, we conduct 5 experiments, namely \textit{Baseline - R}, \textit{Baseline - R + DCA}, \textit{Baseline + DCA}, \textit{Baseline + DCA + DAN-ml(mask loss)} and \textit{Baseline + DCA + DAN-al(affinity loss)}. It’s worth noting that \textit{Baseline - R + DCA} and \textit{Baseline + DCA} both use DCA, but \textit{Baseline + DCA} use radar ranging to alignment the coordinates while \textit{Baseline - R + DCA} use image ranging.

The results are shown in Table \ref{table:ranging_acc}.

\begin{table*}
	\centering
	\begin{tabular}{r|c|c|c|c|c|c}
		\hline
		\multirow{2}{*}{}                & \multicolumn{4}{c|}{Car ranging accuracy}       & \multicolumn{2}{c}{Average ranging accuracy} \\ \cline{2-7} 
		& 0-10m & 10-30m & 30-80m & 80-105m & Car             & CIPV           \\ \hline
		\textit{Baseline - R}            & 0.7238   & 0.4985    & 0.4821    & 0.3411       & 0.4977          & 0.6162         \\ 
		\textit{Baseline - R + DCA}      & 0.8460   & 0.5776    & 0.4776    & 0.2424       & 0.5210          & 0.6661         \\ 
		\textit{Baseline + DCA}          & 0.8809   & 0.7447    & 0.6304    & 0.3646       & 0.6665          & 0.7823         \\ 
		\textit{Baseline + DCA + DAN-ml} & 0.7640   & 0.6407    & 0.4930    & 0.3949       & 0.5519          & 0.6539         \\ 
		\textit{Baseline + DCA + DAN-al}                & \textbf{0.8869}   & \textbf{0.7463}    & \textbf{0.6366}    & \textbf{0.4164}       & \textbf{0.6720}          & \textbf{0.7934}         \\ \hline
	\end{tabular}
	\caption{Ranging accuracy on the test set.}
	\label{table:ranging_acc}
\end{table*}

\textbf{DCA.} As we can see from the TableII, Baseline - R + DCA perform better than Baseline – R. The ranging accuracy of closest in-path vehicle (CIPV) rises $5\%$ and car rises $2.3\%$, which demonstrate the effectiveness of the proposed DCA.

\textbf{DAN.} The DAN trained with affinity loss has the best performance among all the methods. More precisely, the ranging accuracy of CIPV rises $1\%$ and car beyond 80M rises $5\%$, which shows the DAN is better than and can be used to replace traditional artificial strategy.

\textbf{Loss function.} DAN with affinity loss has an obvious performance advantage. When trained with mask loss, the convergence of the network is slower and harder, which makes the ranging accuracy decreases obviously.

Qualitative results of the \textit{Baseline - R} and \textit{Baseline + DCA + DAN-al} are shown in Figure \ref{exp:qualitative}, including flat and rough road, day and night, sunny and rainy days. It is obvious that the proposed multi-modality cascaded framework can improve the ranging accuracy, which means preferable fusion results of camera and radar.

\section{Conclusion}

We propose a multi-modality cascaded fusion framework, supporting various sensors fusion with great interpretability. In addition, the dynamic coordinate alignment can facilitate the feature extraction and can be adapted to other sensor fusion methods. Moreover, the affinity loss function is more suitable for practical applications, which eases the model convergence and improves association accuracy. Finally, the hierarchical association and fusion framework is insensitive to single modality sensor failure, making the entire perception results more robust and can better serving autonomous driving decisions.

{\small
\bibliographystyle{ieee_fullname}
\bibliography{egbib}
}

\end{document}